%% file: main.tex
\documentclass[runningheads]{llncs}
\usepackage[T1]{fontenc}
\usepackage{amsmath,amssymb}
\usepackage{booktabs}
\usepackage{placeins}
\usepackage{xcolor}
\usepackage{hyperref}
\usepackage{microtype}
\usepackage{graphicx}
\author{Benjamin Léger\inst{1} \and
Kazem Meidani\inst{2} \and
Christian Gagné\inst{1,3}}
\authorrunning{F. Author et al.}
\institute{IID / Mila, Université Laval, Québec, Canada  \and
Department of Mechanical Engineering, Carnegie Mellon University
\\
Canada-CIFAR AI Chair}

\providecommand{\Description}[1]{}
\begin{document}
\title{Multi-Modal Learning meets Genetic Programming: Analyzing Alignment in Latent Space Optimization}
%
%\titlerunning{Abbreviated paper title}
% If the paper title is too long for the running head, you can set
% an abbreviated paper title here
%
\titlerunning{Multi-Modal Learning and GP for Latent Space Optimization}

% \author{First Author\inst{1}\orcidID{0000-1111-2222-3333} \and
% Second Author\inst{2,3}\orcidID{1111-2222-3333-4444} \and
% Third Author\inst{3}\orcidID{2222--3333-4444-5555}}
% %
% \authorrunning{F. Author et al.}
% % First names are abbreviated in the running head.
% % If there are more than two authors, 'et al.' is used.
% %
% \institute{Princeton University, Princeton NJ 08544, USA \and
% Springer Heidelberg, Tiergartenstr. 17, 69121 Heidelberg, Germany
% \email{lncs@springer.com}\\
% \url{http://www.springer.com/gp/computer-science/lncs} \and
% ABC Institute, Rupert-Karls-University Heidelberg, Heidelberg, Germany\\
% \email{\{abc,lncs\}@uni-heidelberg.de}}

% \author{Benjamin Léger}\affiliation{\institution{IID / Mila, Université Laval}\city{Quebec City (Quebec)}\country{Canada}}\email{benjamin.leger.1@ulaval.ca}
% \and 
% \author{Kazem Meidani}\affiliation{\institution{Department of Mechanical Engineering, Carnegie Mellon University}}

% \author{Christian Gagné}\affiliation{\institution{Canada-CIFAR AI Chair\\ IID / Mila, Université Laval}\city{Quebec City (Quebec)}\country{Canada}}\email{christian.gagne@gel.ulaval.ca}

%
\maketitle              % typeset the header of the contribution
\begin{abstract}
\input{sections/abstract}
\end{abstract}
% \textcolor{red}{TODO : 
% \begin{itemize}
%     \item Switch to CMA-ES results and modify all refs to CMA-ES in the text 
%     \item Better explain why Cosine-similarity is the good metric for alignemt
%     \item Mention GenSR? explicit difference and why work still relevant (e.g. SNIP has it's own advantages, study 2 claims brooder than just SR?)
%     \item Give CMA-ES parameters 
%     \item Recheck all reviews and see if we've adressed all important things
%     \item Cleaner le repository pour pouvoir le fournir
%     \item If time and space give equations with good and bad alignments 
%     \item Increase size of the figure 1 
% \end{itemize}
% }
\input{sections/introduction}

\input{sections/preliminaries}

\input{sections/experiments}
\input{sections/discussion}

\input{sections/related_works}

\input{sections/conclusion}

% ---- Bibliography ----
%
% BibTeX users should specify bibliography style 'splncs04'.
% References will then be sorted and formatted in the correct style.
%
% \bibliographystyle{splncs04}
% \bibliography{mybibliography}
%
\bibliographystyle{splncs04}
\bibliography{biblio_v0}

\end{document}

%% file: sections/abstract.tex
Symbolic regression (SR) aims to discover mathematical expressions from data, a task traditionally tackled using Genetic Programming (GP) through combinatorial search over symbolic structures. Latent Space Optimization (LSO) methods use neural encoders to map symbolic expressions into continuous spaces, transforming the combinatorial search into continuous optimization. SNIP (Meidani et al., 2024), a contrastive pre-training model inspired by CLIP, advances LSO by introducing a multi-modal approach: aligning symbolic and numeric encoders in a shared latent space to learn the phenotype-genotype mapping, enabling optimization in the numeric space to implicitly guide symbolic search. However, this relies on fine-grained cross-modal alignment, whereas literature on similar models like CLIP reveals that such an alignment is typically coarse-grained. In this paper, we investigate whether SNIP delivers on its promise of effective bi-modal optimization for SR. Our experiments show that: (1) cross-modal alignment does not improve during optimization, even as fitness increases, and (2) the alignment learned by SNIP is too coarse to efficiently conduct principled search in the symbolic space. These findings reveal that while multi-modal LSO holds significant potential for SR, effective alignment-guided optimization remains unrealized in practice, highlighting fine-grained alignment as a critical direction for future work.

%% file: sections/introduction.tex
\section{Introduction}\label{sec:introduction}

% Context + Motivation : 
%% GP is combinatorial, approaches have tried to learn space to make continuous
%% Approach very interesting, but they mostly focus on symbolic encodings 
While traditional Genetic Programming typically explores vast symbolic solution spaces through combinatorial search, recent methods \cite{meznar2023efficient,liskowski2020program,kusner2017grammar,caetano2023symbolic} -- referred to here as Latent Space Optimization Genetic Programming, or LSO-GP -- transform this combinatorial problem into continuous optimization. Leveraging the success of deep representation learning methods, these approaches use neural encoders to map symbolic expressions into continuous and semantically dense latent spaces. Traditional black-box optimizers are subsequently used to conduct the search over this space. However, most existing LSO-GP methods focus exclusively on encoding the \emph{symbolic} structure (i.e., genotype) of candidate solutions, without capturing information about their numeric behavior (i.e. phenotype or semantics)..
%% --> SR is by definition a multi modal task --> two types of objectives --> ideally we want an algorithme to optimize in both spaces and include both informations 
Symbolic Regression (SR) is, however, a multi-modal task \cite{li2024mmsr,meidani2024snip}, as it aims to find interpretable \emph{symbolic} equations from \emph{numeric} data. Understanding the underlying \emph{genotype-phenotype mapping}, that is, data-fitting accuracy, i.e., how well the solution fits the data, has long been recognized as key for the design of effective SR heuristics \cite{winkler2018similarity}. The field of Semantic Genetic Programming \cite{vanneschi2014survey} has demonstrated how semantic information can guide symbolic search more efficiently. Moreover, SR inherently requires optimizing two distinct types of objectives: numerical accuracy, or how well the solution fits the data, and symbolic relevance, which includes interpretability and recovery of meaningful syntactic patterns \cite{bertschinger2024evolving,yu2024mdlformer}. Rather than treating these modalities separately, exploiting their interactions and mutual information is crucial for designing effective search algorithms \cite{li2024mmsr}.

Recent advances in multi-modal models offer a potential solution to this challenge. Vision-Language Models (VLMs) like CLIP \cite{radford2021learning} can map complex data to information-rich continuous spaces while learning relationships between different modalities of the same concept. This capability suggests multi-modal models could provide the missing ingredient for data-driven Genetic Programming: algorithms that explore continuous spaces which are simultaneously informative about both the numeric and symbolic nature of candidate solutions. Building on this idea, SNIP \cite{meidani2024snip}, inspired by CLIP, uses a pretrained bi-modal model to encode both numeric and symbolic modalities of mathematical equations. The model is trained to align these representations by mapping both modalities of the same equation close to each other in a shared latent space. For symbolic regression, SNIP conducts LSO search in the numeric latent space, with the premise that the learned inter-modal alignment will implicitly guide exploration of the symbolic space.
% Problem 
%% The fact it works depends mostly on TWO things : (1) does the algorithm really use that (after all it only explores on space) --> (2) if it does it's only based on the hypothesis that alignment is very fine grained (that would be necessary, and ANYWAY necessary to have LSO-GP methods that are good with semantics). 
%% --> Litterature on CLIP, from which it's derived shows its not the case 
% But hopes for multi-modal optimization and simultaneous symbolic-numeric awareness promises depens on two main key success factors. On the \emph{search algorithm side}, it is not obvious that the proposed strategy effectively \emph{exploits} the said alignment. 
% On the \emph{model side}, we argue that, in order to have the ability to efficiently and accurately evolve and improve symbolic forms of solutions over the course of optimization, it is necessary for the model to possess an acute understanding of what subtle symbolic modications and alterations mean, and what they imply in term of numerical behavrioral changes. However, the litterature on similar bi-modal contrastive model reveal their unability to capture and understand fine semantic subtlities, raising doubts on SNIP's default ability on the matter \cite{lewis2022does, wang2023equivariant, tong2023mass, chen2025understanding}.
However, the effectiveness of this approach depends on two critical assumptions. First, the LSO algorithm must actively exploit the learned alignment during search. Second, the alignment itself must be fine-grained enough to distinguish between symbolically different expressions. Existing literature on contrastive bi-modal models like CLIP reveals systematic failures to capture fine-grained semantic distinctions \cite{lewis2022does,wang2023equivariant,tong2023mass,chen2025understanding}, raising the question of whether SNIP's alignment is sufficiently fine-grained for effective symbolic optimization.

% Problem definition and contributions 
%% We analyse both success factor for this goal for semantic aware LSO-optimization, and we show that they are not met. 
% In this work, we analyze both success factors and confirm the stated doubt. More specifically our contributions are the following : 
% \begin{enumerate}
%     \item We empirically outline the current lack of exploitation of SNIP-LSO learned alignment
%     \item We bridge the gap with the contrastive learning literature and evaluate the effective cross-modal alignment granularity of SNIP, revealing it not to be fine-grained enough for truly effective symbolic optimization 
%     \item We outline and discuss a clear path for improving Latent Space Optimization methods for GP, and imagine other possible applications of these models outside the strict scope of LSO.
% \end{enumerate}
In this work, we investigate these two assumptions to better understand the strengths and limitations of multi-modal LSO for symbolic regression. Specifically, we make the following contributions:
\begin{enumerate}
\item We empirically demonstrate that SNIP's current LSO formulation does not actively exploit the learned cross-modal alignment during optimization, even as fitness improves.
\item We evaluate the granularity of SNIP's cross-modal alignment using retrieval tasks adapted from the contrastive learning literature, revealing that the alignment is too coarse to reliably distinguish between structurally similar symbolic expressions.
\item We discuss implications for future work, identifying fine-grained alignment as a critical direction for improving multi-modal LSO methods and outlining potential paths forward for symbolic regression.
\end{enumerate}

% Contributions 
%% 1- Study and outline the current lack of exploitation of SNIP-LSO of learned alignment
%% 2- We evaluate the alignment granularity of the model, showing it is not fine-grained enough for symbolic regression, (2.1) at the same time bridge the gap with litterature on SNIP
%% 3- Outline and discuss a clear path for improving Latent Space Optimization methods for GP 

% \textcolor{red}{
% \begin{itemize}
%     \item Message probably still a bit unclear.
%     \item Perhaps not high level enough, maybe doesn't sell the point enough (perhaps borough ideas from the discussion, e.g. end goal of learning data drive GP algorithms that generalize over all concepts studied over years by the GP community : phenotype genotype mapping, mapping to continuous optimzation, semantic awareness and semantic/behavioral GP. Like "from heuristic implemtantions of these concepts to an automatic generalized version of it.)
%     \item Perhaps a little bit more focus and emphasis of the general potential of Multi-Modal models for GP and SR in general.
% \end{itemize}
%  }

%% file: sections/preliminaries.tex
\begin{figure}[t]
  \includegraphics[width=\linewidth,height=0.64\textheight,keepaspectratio]{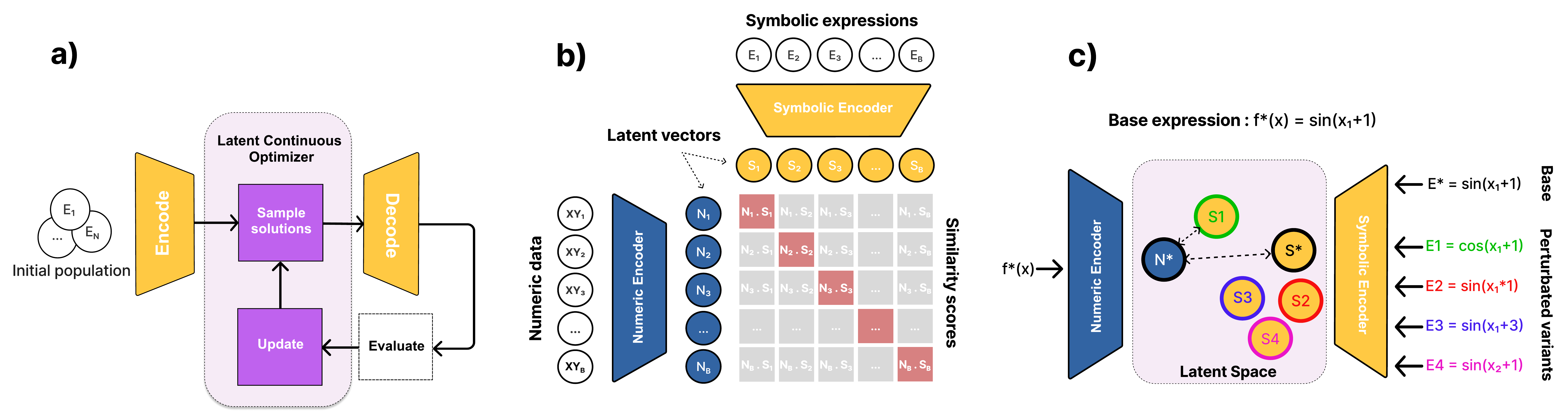}
  \caption{Overview of Latent Space Optimization and multi-modal alignment. a) Traditional LSO-GP framework : a continuous optimizer searches the latent space of a symbolic autoencoder. b) SNIP's \cite{meidani2024snip} contrastive pre-training: symbolic and numerical encoders are trained to align embeddings of matched expression pairs with batch. c) Illustration of alignment granularity limitations : given a base expression $f^{\star} = \sin(x+1)$ and its numeric embedding $Z_V^{\star}$ (labeled $N^{\star}$ in the figure) encoded from observations $(X,y^{\star})$, the model should assign highest similarity between $Z_V^{\star}$ and the symbolic embedding $Z_S^{\star}$ (labeled $S^{\star}$) over embeddings of perturbed variants ($S_1, S_2, S_3$). In practice, the learned alignment often fails to make this distinction.}
  \Description{TBD.}
  \label{fig:preliminaries}
\end{figure}

% Previous bullet point version 
\section{Preliminaries and background}\label{sec:preliminaries}

\subsection{Latent Space Optimization for SR}\label{sec:preliminaries-LSOGP}
Latent Space Optimization (LSO) transforms the combinatorial search in Genetic Programming (GP) algorithms into continuous optimization in learned representation spaces. The typical approach involves training an autoencoder to map symbolic expressions to continuous embeddings and reconstruct them via a decoder, minimizing syntactic reconstruction objectives such as cross-entropy loss. Once this space is constructed, gradient-free or gradient-based optimization algorithms can be utilized to search and sample candidates from this space, which are then decoded into symbolic expressions and evaluated for their fitness on the target data (as depicted in Fig~\ref{fig:preliminaries}-a). Prior LSO methods for symbolic regression \cite{caetano2023symbolic,kusner2017grammar,meznar2023efficient} primarily encode symbolic structure without explicitly representing numeric semantics.

\subsection{SNIP : Multi-modal Pretraining for Mathematical Equations } \label{sec:preliminaries-SNIP} % \textcolor{red}{[Some part generated with help of AI]}

% Based on the success of popular Visual Language Models (VLM) such as CLIP \cite{radford2021learning}, SNIP \cite{meidani2024snip} introduces a new direction for this field. The paper proposes a multi-modal pre-training that jointly learns symbolic and numeric representations of mathematical expressions. With this aim to learn complexe relationships between both modalities, related to the well known Genetopy-Phenotype mapping, that type of method opens intriguing perspective for the field of Symbolic Regression where the joint exploitation of both informations is known to be a key success factor \cite{}. In particular the authors propose a LSO approach to explore the learned spaces. We give here an overview of the method and discuss key points related to its use in Symbolic Regression and Genetic Programming.
SNIP~\cite{meidani2024snip} is a multi-modal pre-training framework that jointly learns symbolic and numeric representations of mathematical expressions. Inspired by CLIP~\cite{radford2021learning}, SNIP uses contrastive learning to align both modalities in a shared latent space, aiming to capture the genotype-phenotype mapping between symbolic structure and numeric behavior. We provide an overview of SNIP's architecture, training procedure, and its application to symbolic regression through latent space optimization.

\subsubsection{Architecture and training}\label{sec:preliminaries-SNIP-archi}
SNIP's architecture consists of two transformer encoders: one processing symbolic expressions (as prefix-order token sequences), another processing numeric observations (input-output pairs $(x, y)$), resulting in latent vector representations for each modality, respectively denoted as $Z_S$ (for symbolic embeddings) and $Z_V$ (for numeric embeddings) in the rest of the paper. Both encoders are trained with the same contrastive objective used in CLIP, the \textit{InfoNCE} loss \cite{oord2018representation}, learning to align embeddings of matched symbolic-numeric pairs while separating unrelated pairs (as illustrated in Fig~\ref{fig:preliminaries}-b): 

% \resizebox{\columnwidth}{!}{\label{eq:infonce}
% \begin{equation}
%     InfoNCE = -\sum_{(s,v) \in \mathcal{B}} \left[
%       \log \frac{\exp(Z_S \cdot Z_V^+ / \tau)}{\sum_{Z \in \{Z_V^+, Z_V^-\}} \exp(Z_S \cdot Z / \tau)}
%       + \log \frac{\exp(Z_V \cdot Z_S^+ / \tau)}{\sum_{Z \in \{Z_S^+, Z_S^-\}} \exp(Z_V \cdot Z / \tau)}
%     \right]
% \end{equation}}

% \begin{equation}
%   \mathcal{L} = -\log \frac{\exp(\text{sim}(Z_V, Z_S^+) / \tau)}{\sum_{i} \exp(\text{sim}(Z_V, Z_{S,i}) / \tau)}
%   \label{eq:infonce}
% \end{equation}
% where $Z_V$ is the anchor representation, $Z_S^+$ is the positive sample, $Z_{S,i}$ represents all samples in the batch including the negatives, and $\tau$ is the temperature scalar.

\begin{equation}\label{eq:infonce}
\begin{aligned}
    \mathcal{L}_{\text{InfoNCE}} = -\sum_{(S,V) \in \mathcal{B}} \bigg[
      &\log \frac{\exp(Z_S \cdot Z_V^+ / \tau)}{\sum_{Z \in \{Z_V^+, Z_V^-\}} \exp(Z_S \cdot Z / \tau)} \\
      &+ \log \frac{\exp(Z_V \cdot Z_S^+ / \tau)}{\sum_{Z \in \{Z_S^+, Z_S^-\}} \exp(Z_V \cdot Z / \tau)}
    \bigg]
\end{aligned}
\end{equation}

where $\mathcal{B}$ is a minibatch of randomly generated mathematical equations, $\tau$ a learnable temperature coefficient, and  $Z_{S}^{+}$, $Z_{V}^{+}$, $Z_{S}^{-}$ and $Z_{V}^{-}$ denote positive (matched) and negative (unmatched) pairs for each modality. Concretely, for a given expression $f$ with symbolic form $S$ and numeric observations $V$, the positive pair is $(Z_S, Z_V)$ from the same expression, while negative pairs are formed by all other expressions in the batch. This formulation encourages a form of geometric coherence between the two modalities in a ``joint'' learned space, with the main objective of training models that capture the underlying relationship between them.

% \subsubsection{Inter-modal alignement}\label{sec:preliminaries-SNIP-align-def} \textcolor{red}{[SOME WORK TO DO ON BEING PRECISE AND CLEAR ABOUT THE TERMS]}
% We refer in this paper to the concept of  \textit{inter-modal alignment} broadly as \textit{how well the model associates related concepts across modalities in the learned latent space}". If we mainly refer to it as a general sense, we will pay specific attention in our experiments at how "close" two embeddings $Z_V, Z_S$ of two different modalities are in the learned latent space, using for instance their cosine similarity: 
% \begin{equation}
% \cos\left(Z_S, Z_V\right)
% \end{equation}

% The training procedure described in section \ref{sec:preliminaries-SNIP-archi} and the Info-NCE loss described in equation \ref{eq:infonce} explicitely maximises this quantity for two modalities of a same expression relatively to modalities of others. However, we'll consider in section \ref{sec:experiments-study1} the alignement between modalities of different expressions as a metric for symbolic optimization. 
% \km{We can move this subsection completely to next section where we discuss alignment problem -- it's not that much about background}

\subsubsection{Latent Space Optimization for Symbolic Regression}
SNIP applies LSO to symbolic regression in a framework similar to 
the general approach described in Section~\ref{sec:preliminaries-LSOGP}, with a key 
difference: the search is conducted in the continuous latent space 
produced by the \emph{numeric} encoder, rather than a symbolic one. 
A decoder network, inherited and fine-tuned from prior neuro-generative 
SR work~\cite{kamienny2022end}, maps latent vectors to symbolic 
expressions. Given a target dataset $(X, y^\star)$, the optimization 
procedure iteratively searches for a latent vector $z \in \mathbb{R}^d$ 
whose decoded expression maximizes data-fitting accuracy on the target:

\begin{enumerate}
    \item \textbf{Target encoding.} The numeric encoder produces the 
    target latent vector $Z_V^\star = \mathrm{encode}_V(X, y^\star)$.
    
    \item \textbf{Initialization.} A population of $P$ latent vectors 
    is initialized around $Z_V^\star$ through dataset augmentation and 
    latent perturbations (details in Section~\ref{sec:experiments-study1}).
    
    \item \textbf{Sampling.} At each iteration, a black-box optimizer 
    (e.g., CMA-ES~\cite{hansen2023cmaevolutionstrategytutorial}, GWO, or any other population-based 
    search algorithm) samples new candidate latent vectors from its 
    current search distribution.
    
    \item \textbf{Decoding and refinement.} Each candidate $z$ is decoded 
    into a symbolic expression $f_z$; its numeric constants are then 
    refined with BFGS~\cite{fletcher2013practical} on $(X, y^\star)$.
    
    \item \textbf{Evaluation and update.} Each $f_z$ is scored by its 
    $R^2$ fitness on $(X, y^\star)$. The optimizer uses these fitness 
    values to update its sampling distribution for the next iteration 
    (step 3), biasing the search toward high-fitness regions of the 
    latent space.
\end{enumerate}

\section{Alignment in Multi-Modal LSO}\label{sec:motivation}

\subsection{SNIP's Performance: Promise and Gaps}\label{sec:motivation-snip-gaps}

SNIP demonstrates promising capabilities for symbolic regression. The approach shows strong data-fitting accuracy and generates solutions with reasonable complexity on the SRBench benchmark \cite{la2021contemporary}. Evidence suggests the model learns meaningful cross-modal relationships: LSO in the learned space yields better results than single-shot predictions decoded directly from the target's numeric embedding, the model can quickly classify mathematical properties of expressions directly from their symbolic form, and visualizations reveal meaningful clusters of basic symbolic properties (e.g., number of variables, operator types) in the numeric space and vice versa (e.g., convexity, monotonicity in the symbolic space).

However, the key hypothesis underlying SNIP's multi-modal approach, that progressive moves in the  continuous latent space translate to meaningful symbolic modifications via learned alignment, remains unvalidated. Recent work by Yu et al. \cite{yu2024mdlformer} demonstrates that despite SNIP's data-fitting performance, it struggles to effectively retrieve relevant symbolic forms, showing significantly inferior symbolic retrieval rates compared to GP-based heuristics. Notably, SNIP fails to show substantial improvement over previous transformer-based generative models that include no explicit symbolic constraints in their search. This gap between data-fitting performance and symbolic retrieval raises fundamental questions about whether the learned alignment effectively guides symbolic search during optimization.

\subsection{Limitations of Contrastive alignment and CLIP-like multi-modal models}
SNIP's training objective mirrors CLIP's \cite{radford2021learning}, which learns joint vision-language representations through contrastive learning. The extensive literature analyzing CLIP reveals systematic limitations that may be relevant for understanding SNIP's performance. While CLIP achieves impressive zero-shot performance on many tasks, its alignment captures high-level semantic similarity rather than fine-grained structure. Studies show CLIP behaves as a ``bag of concepts'', failing to bind attributes to objects or distinguish compositional relationships \cite{lewis2022does}. Similarity scores do not vary faithfully with semantic changes \cite{wang2023equivariant}, and the model struggles with quantifiers, negations, and spatial relations \cite{tong2023mass,chen2025understanding}. For SNIP, analogous limitations could mean the model captures coarse numerical behavior but fails to distinguish symbolically different expressions with similar outputs.

These findings are particularly concerning for symbolic regression. Effective LSO in the symbolic space requires the ability to model how small symbolic modifications translate into behavioral changes. If CLIP's contrastive alignment fails on fine-grained visual-linguistic distinctions despite its scale and success, SNIP's alignment may similarly struggle to capture the precise symbolic-numeric relationships needed for principled symbolic search.

\subsection{Measuring Cross-Modal Alignment}

We refer to \textit{inter-modal alignment} as how well the model associates the symbolic form of an expression with its numeric behavior (the two \emph{modalities}) in the learned latent space. We quantify alignment using cosine similarity between embeddings. This is a natural choice: it is directly optimized by the InfoNCE objective (Eq. 1), and is the standard similarity measure in contrastive representation learning:
% \begin{equation}
% \cos\left(Z_S, Z_V\right)
% \end{equation}
\begin{equation}
\text{alignment} = a(Z_S, Z_V) = \cos(Z_S, Z_V) = \frac{Z_S \cdot Z_V}{\|Z_S\| \|Z_V\|}
\label{eq:alignment}.
\end{equation}
The InfoNCE loss (Eq.~\ref{eq:infonce}) explicitly maximizes this similarity for matched symbolic-numeric pairs of the same expression during training while minimizing it for unmatched pairs. In principle, this learned alignment could guide the optimization, that is if a candidate solution's symbolic embedding $Z_S$ has high alignment with the target's numeric embedding $Z_V^*$, this could indicate symbolic relevance to the problem. However, whether the alignment is actually exploited during LSO and whether it is fine-grained enough to support symbolic search remains an open question.

\subsection{Research Hypotheses } % \textcolor{red}{Or Research Questions? we can frame as Questions too}

Given the gap between SNIP's data-fitting performance and its symbolic retrieval performance, combined with known limitations of contrastive alignment in similar models, we investigate two hypotheses for why the learned alignment may not effectively guide symbolic search:
\begin{description}
\item[H1 -- Algorithm exploitation:] The LSO procedure does not actively exploit the learned cross-modal alignment during optimization, even if the alignment quality is sufficient.
\item[H2 -- Alignment granularity:] The learned alignment is too coarse-grained to distinguish between symbolically different expressions, preventing effective symbolic guidance even if the algorithm attempts to exploit it.
\end{description}
These hypotheses are not mutually exclusive. Both could contribute to the observed gap in symbolic performance. In the following sections, we design experiments to test each hypothesis empirically.

%% file: sections/experiments.tex
\section{Experimental Analysis} \label{sec:experiments}
To address the hypotheses outlined above, we pose two research questions : (a) Does SNIP's LSO actively exploit the learned alignment, or does optimization occur independently of cross-modal relationships? (Verifying H1), and (b) Is the learned alignment fine-grained enough to support symbolic search, or does it only capture coarse behavioral similarities? (Verifying H2). We answer these questions through two complementary studies.

\subsection{Study 1: Cross-Modal Alignment During Optimization}\label{sec:experiments-study1}
To verify H1, we investigate whether SNIP actively exploits cross-modal alignment during optimization. We run SNIP's original LSO algorithm and track the following quantities for the best individual at each iteration $t$: (i) $R^2_t$ fitness, indicating whether evolution improves data-fitting accuracy and (ii) Inter-modal alignment between its symbolic embedding and the target's numeric embedding (Eq.~\ref{eq:alignment}). While LSO gradient-free optimizer's is expected to monotonically increases $R^2_t$, the evolution of $a_t$ reveals whether the model's search is guided by symbolic relevance, that is, how symbolically appropriate the current candidate solution is relative to the symbolic regression target. If the algorithm actively exploits the learned alignment to incorporate symbolic awareness, this quantity should progressively increase over the course of optimization.

\subsubsection{Experimental details.}
We replicate SNIP's experimental setup, hyperparameters, and model parameters (for full details, see \cite{meidani2024snip}), running optimization for $80$ iterations without early stopping to enable consistent comparison of metrics across the entire optimization trajectory. % \textcolor{blue}{We use the CMA-ES algorithm as black-box optmizer.}

\paragraph{Dataset.} We use the Feynman and Strogatz benchmark suites included in the popular symbolic regression benchmark SRBench \cite{la2021contemporary}, keeping all equations with input dimensions $D \leq 10$.

\paragraph{Encoders.} Symbolic and numeric embeddings used to compute alignment were obtained using SNIP's original pre-trained encoders (not fine-tuned for LSO), as provided by the authors. % [TODO: Confirm encoder normalization strategy used.]

\paragraph{Initialization and first measurement.} Following SNIP's LSO procedure, the population is initialized via dataset augmentation: subsampling ($p_1=15$ individuals), target perturbation ($p_2=10$ individuals), and latent perturbation ($p_3=25$ individuals), totaling $P=50$ individuals. The first measurement point ($t=0$) corresponds to the model's one-shot prediction from the target data encoding. Measurement $t=1$ corresponds to the best individual after initialization and a BFGS constant refinement step. Measurement $t=2$ onwards correspond to the best individual after each optimization iteration.

\subsubsection{Results.}

\begin{figure}[t]
  \centering
  \includegraphics[width=0.48\linewidth]{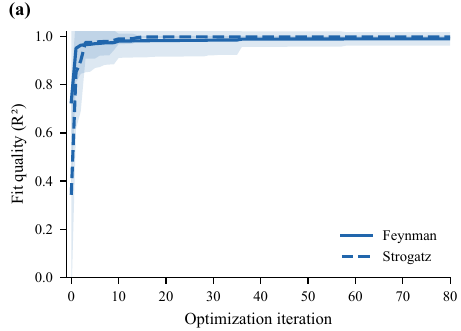}
  \hfill
  \includegraphics[width=0.48\linewidth]{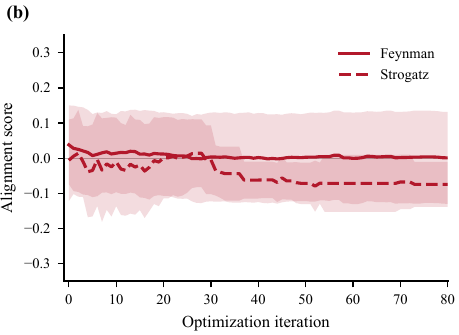}
  \caption{Evolution of (a) $R^2$ fitness (left) and (b) cross-modal alignment (right) during LSO across all Feynman and Strogatz equations, averaged over all equations.  While fitness improves consistently, alignment remains flat or decreases, indicating that optimization does not exploit cross-modal alignment to produce better symbolic solutions. Iteration 0 corresponds to the model's one-shot prediction, iteration 1 to the best individual after initialization and BFGS constant refinement.}
  \label{fig:study1-trajectories}
\end{figure}

Evolution of both metrics is reported in Fig.~\ref{fig:study1-trajectories}. $R^2$ shows the expected increasing trend, improving from $0.73$ at $t=0$ (one-shot prediction) to $0.96$ at $t=1$ (after initialization and BFGS) and reaching $0.99$ at $t=80$ (final iteration) on Feynman. The substantial $R^2$ improvement between $t=0$ and $t=1$ reveals the strong influence of initialization and constant refinement on data-fitting accuracy.
In contrast to numeric accuracy, alignment remains essentially flat throughout optimization, starting at $0.038$ and decreasing slight\-ly to $0.001$ by the final iteration on the Feynman set. These results demonstrate that the search algorithm does not actively exploit the learned alignment during optimization. Even as numerical fitness steadily improves, alignment quality neither increases nor guides the search toward symbolically relevant solutions. This indicates that the LSO procedure operates independently of the cross-modal relationships learned during pretraining. Similar conclusions can be drawn with Strogatz equations. This confirms H1: SNIP's LSO procedure does not actively exploit the learned cross-modal alignment during optimization.

\FloatBarrier

\subsection{Study 2: Alignment Granularity }\label{sec:experiments-study2} % \textcolor{red}{NOTE : Parts of text generated with help of AI}

We now address H2, by verifying whether the learned alignment is fine-grained enough to support symbolic search. While SNIP shows evidence of some cross-modal understanding (e.g., property prediction, operator clustering), effective symbolic optimization may require much finer granularity to distinguish between structurally similar expressions.

We evaluate whether SNIP's learned alignment can discriminate between structurally similar symbolic expressions through a retrieval task. Given a base expression $f$ and its corresponding numerical observations $(X, y)$, we generate $K-1$ perturbed variants $\{f_1', f_2', \ldots, f_{K-1}'\}$ that differ minimally from $f$ in symbolic structure. The task is to identify the correct expression $f$ among the $K$ candidates using only the alignment (cosine similarity, Eq.~\ref{eq:alignment}) between the symbolic embedding $Z_S$ and the numerical embedding $Z_V^*$ encoded from $(X, y)$.

\subsubsection{Experimental details.} % \textcolor{red}{[probably a bit lenghty, things to probably include in appendices]}

\paragraph{Dataset Construction. }

We evaluate on two datasets. First, the Feynman physics equations from SRBench with $2 \leq d \leq 10$ variables. The lower bound ensures applicability of variable substitution perturbations (which require at least two distinct variables), while the upper bound matches SNIP's maximum capacity. After filtering equations without known ground-truth expressions and excluding 12 equations due to numerical issues or parsing failures, this yields 82 test cases. Second, $100$ synthetic expressions generated using SNIP's training data distribution. As these match the pre-training data, they might be seen as an upper bound on expected performance.

\paragraph{Perturbation Strategies.}
For each base equation, we apply up to four symbolic perturbation types designed to create structurally similar but semantically different expressions, summarized in Table~\ref{tab:perturbations}.
\begin{table}[t]
\centering
\caption{Perturbation types used to generate candidate expressions. Each perturbation modifies a single element of the expression tree while preserving overall structure.}
\label{tab:perturbations}
% \resizebox{\columnwidth}{!}{%
\begin{tabular}{lll}
\toprule
Type & Description & Examples \\
\midrule
Unary operator swap & Replace unary operator with & $\sin \leftrightarrow \cos \leftrightarrow \tan$ \\
                    & semantically related alternative & $\exp \leftrightarrow \log$; \quad $\sqrt{\cdot} \leftrightarrow (\cdot)^2 \leftrightarrow (\cdot)^3$ \\[2pt]
Binary operator swap & Replace binary operator with & $+ \leftrightarrow -$; \quad $\times \leftrightarrow \div$ \\
                     & any of the other three & $+ \leftrightarrow \times$; \quad $- \leftrightarrow \div$ \\[2pt]
Constant change & Modify a numeric constant & $2 \to 3$; \quad $\pi \to e$; \quad $c \to 1.2c$ \\[2pt]
Variable substitution & Replace one variable with another & $x_0 \to x_1$ \\
\bottomrule
\end{tabular}%
% }
\end{table}
For unary operators, we group operators into semantic families and allow swaps only within families: trigonometric ($\sin$, $\cos$, $\tan$), exponential/logarithmic ($\exp$, $\log$), and power functions ($\sqrt{\cdot}$, $(\cdot)^2$, $(\cdot)^3$).
For binary operators, we allow swaps between any pair of the four arithmetic operators, covering both within-family swaps (e.g. $+ \leftrightarrow -$, $\times \leftrightarrow \div$) and cross-family swaps (e.g. $+ \leftrightarrow \times$, $- \leftrightarrow \div$).
When multiple instances of a swappable operator exist in an expression, one is selected uniformly at random. As not all perturbations are applicable to every equation (e.g., unary operator swap requires the presence of a swappable unary operator), each test case includes in average $\bar{K} = 3.67$ candidates (1 base + 2.67 perturbations) for the Feynman set and $\bar{K} = 4.42$ for the synthetic set. 

\paragraph{Evaluation Protocol.}
For each test case, we:
\begin{enumerate}
    \item Load data $(X, y)$ with $N=200$ standardized input points;
    \item Encode the numeric target: $Z_V^* = \text{encode}_V(X, y)$ using SNIP's original pre-trained encoder;
    \item Encode all $K$ candidate expressions: $Z_S^{(k)} = \text{encode}_S(f_k)$ using the LSO-finetuned encoder;
    \item Rank candidates by alignment $a(f_k) = \cos(Z_S^{(k)}, Z_V^*)$;
    \item Record the rank of the true expression $f^*$.
\end{enumerate}

\paragraph{Metrics.}
We report retrieval accuracy: the fraction of test cases where the base expression ranks first. We compare against a random baseline corresponding to random selection over the $K$ valid candidates for each given evaluation case.

\subsubsection{Results.}

Table~\ref{tab:study2_overall} presents retrieval performance on both data\-sets. On Feynman, the model achieves $18.3\%$ accuracy, below the random baseline of $27.3$ ($0.67$× baseline). This indicates that the alignment signal not only fails to discriminate between similar expressions, but actively misleads the ranking: perturbed expressions often have higher similarity to the target than the correct expression. On Synthetic data, accuracy matches the random baseline exactly ($23.0\% $vs. $22.9\%$), meaning that the model performs no better than chance even on in-distribution expressions. Figure 3 shows the rank distribution for Feynman: the correct expression ranks first in only $15$ cases ($18.3\%$), while it most frequently ranks second ($47.6\%$).\looseness=-1

\begin{table}[t]
  \centering
  \begin{minipage}[t]{0.46\textwidth}
    \centering
    \caption{Overall retrieval performance on both datasets. The model performs worse than random chance on Feynman, not better than random chance on synthetic data.}
    \label{tab:study2_overall}
    \resizebox{\linewidth}{!}{%
    \begin{tabular}{lcc}
    \toprule
    Metric & Feynman & Synthetic \\
    \midrule
    Test cases & 82 & 100 \\
    Retrieval accuracy & 18.3\% & 23.0\% \\
    Random baseline & 27.3\% & 22.9\% \\
    Accuracy / Baseline & $0.67\times$ & $1.00\times$ \\
    Mean rank & $2.29 \pm 0.92$ & $2.13 \pm 0.85$ \\
    \bottomrule
    \end{tabular}%
    }
  \end{minipage}
  \hfill
  \begin{minipage}[t]{0.50\textwidth}
    \centering
    \caption{Per-perturbation confusion analysis. ``Fooling rate'' indicates how often each perturbation type achieves higher similarity than the correct expression.}
    \label{tab:study2_perturbation_results}
    \resizebox{\linewidth}{!}{%
    \begin{tabular}{lccc}
    \toprule
    Perturbation Type & Occurrences & Fooling Rate & Wins \\
    \midrule
    Unary op. swap & 18 & 55.6\% & 5 \\
    Binary op. swap & 79 & 65.8\% & 33 \\
    Constant change & 49 & 49.0\% & 13 \\
    Variable sub. & 82 & 24.4\% & 16 \\
    \bottomrule
    \end{tabular}%
    }
  \end{minipage}
\end{table}
\begin{figure}[t]
  \centering
  \includegraphics[width=0.55\linewidth]{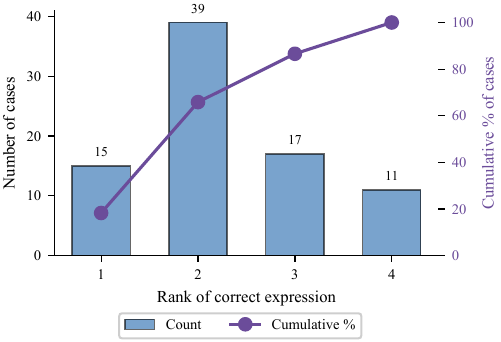}
  \caption{Distribution of ranks for the correct expression across 82 test cases of Feynman set. The correct expression most frequently ranks second (39 cases), with only 15 cases achieving rank 1.}
  \label{fig:study2-ranks}
\end{figure}

\paragraph{Per-Perturbation Analysis.}
Table~\ref{tab:study2_perturbation_results} shows which perturbation types most frequently ``fool'' the model on Feynman. Binary operator swaps are most problematic, fooling the model $65.8\%$ of the time and accounting for $49\%$ of all ranking failures. This suggests the alignment captures general functional form but cannot distinguish expressions that differ only in arithmetic operations. Variable substitutions are easiest to detect (only $24.4\%$ fooling rate), indicating the model does encode some variable-specific information. This last observation is intuitively unsurprising: during pretraining, the numeric encoder receives observations ordered by variable index (i.e. as a stacked tensor $[X_0, X_1,\ldots,X_{D}]$) while the symbolic encoder processes variable through similarily annoted tokens ($x_0,x_1,\ldots,x_{d}$), making variable identity a straightforward correspondence to learn. Per-perturbation patterns are consistent on synthetic data: binary operator swaps remain most challenging ($50.0\%$ fooling rate) and variable substitutions easiest to detect ($12.0\%$ fooling rate).\looseness=-1

\begin{figure}[t]
\centering
\includegraphics[width=0.9\textwidth]{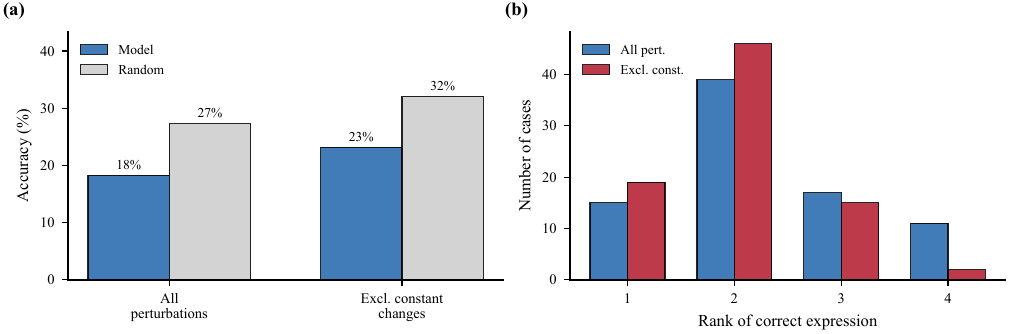}
\caption{Sensitivity analysis excluding constant change perturbations on Feynman. \textbf{(a)} Accuracy comparison showing that even without constant changes, model accuracy (23.2\%) remains below random baseline (32.1\%). \textbf{(b)} Rank distribution comparison showing improved but still suboptimal performance when constant changes are excluded.}
\label{fig:sensitivity}
\end{figure}

\paragraph{Impact of constant swaps.}
Since the handling and representations of constants have been shown to be challenging for transformer-based equation encoders \cite{kamienny2022end,li2022transformer}, and since SNIP performs a constant optimization step after each iteration, we repeat the experiment excluding constant perturbations entirely. Fig.~\ref{fig:sensitivity} shows the impact of removing the ``constant change'' category on the overall results on the Feynman set.
While excluding it improves accuracy from 18.3\% to 23.2\%, confirming that constant perturbations are indeed challenging, accuracy still falls short of the random baseline (23.2\% vs 32.1\%), demonstrating that the alignment's limited discriminative power is \emph{not} solely attributable to constant handling. The model also struggles with operator swaps, confirming that the coarse alignment granularity is a fundamental characteristic rather than a constant-specific artifact. Similar conclusions are noted with the Synthetic set. \looseness=-1

%% file: sections/discussion.tex
%----------------------------------------------------------
\section{Discussion}\label{sec:discussion}

\subsection{Summary of Findings}

Our experimental investigation addressed two research questions about multi-modal latent space optimization for symbolic regression. Study 1 (Sec.~\ref{sec:experiments-study1}) demonstrated that SNIP's LSO procedure does not actively exploit the learned cross-modal alignment during optimization. Despite steady improvements in numerical fitness ($R^2$), alignment between candidate solutions and the target remains flat or decreases throughout the search process. This reveals that the algorithm operates independently of the cross-modal relationships learned during pretraining.

Study 2 (Sec.~\ref{sec:experiments-study2}) evaluated whether the learned alignment is sufficiently fine-grained to support symbolic search. Through retrieval tasks with structurally similar expressions, we found that SNIP's alignment is too coarse-grained to reliably distinguish between expressions that differ in operators or constants. The model achieves only 18.3\% retrieval accuracy, below the 27.3\% random baseline, indicating that perturbed expressions often align more strongly with targets than correct expressions. The model's inability to differentiate solutions with potentially significant behavioral differences when their symbolic forms are close rules out the possibility of using alignment to guide symbolic space optimization.

The combination of Studies 1 and 2 reveals a fundamental gap: alignment neither increases during optimization (Study 1), nor would such increases guide symbolic search effectively if they occurred (Study 2). This explains the symbolic retrieval gap observed by Yu et al. \cite{yu2024mdlformer}. Furthermore, these results show that using alignment as an explicit objective would be bottlenecked by the model's limited discriminative capacity: even if alignment increased during evolution, this would likely not result in meaningful symbolic improvements.

Despite these limitations, our analysis is constructive: by isolating two precise bottlenecks---algorithmic exploitation and alignment granularity---it provides a concrete roadmap for improving multi-modal LSO methods. The potential of such methods remains significant: learning the phenotype-genotype mapping in a continuous, optimizable space could shift SR from hand-crafted heuristics to data-driven search paradigms. Realizing this potential requires addressing the specific challenges identified here.

\subsection{Implications for Multi-Modal LSO}

These results have broader implications beyond SNIP. They reveal a fundamental challenge for multi-modal latent space optimization methods in symbolic regression: learning cross-modal alignment that is simultaneously robust enough to emerge from contrastive pretraining and fine-grained enough to support symbolic search. The success of CLIP in vision-language tasks might suggest that similar architectures and objectives would naturally transfer to symbolic-numeric domains. Our findings demonstrate otherwise.

The distinction between coarse-grained semantic alignment and fine-grained structural alignment is critical. In vision-language tasks, CLIP's ability to capture high-level semantic similarity (e.g., ``a dog playing in a park'') suffices for many applications. In symbolic regression, however, effective search requires understanding how minimal symbolic modifications (e.g., $\sin(x) \to \cos(x)$ or $x + y \to x \times y$) translate to behavioral changes. This demands a qualitatively different level of alignment granularity.

For the broader LSO-GP community, this suggests that incorporating numeric information into latent spaces requires more than simply adding a numeric encoder and applying contrastive learning. The alignment must be explicitly designed and evaluated for the fine-grained discriminative capacity that symbolic search demands. This represents both a challenge and an opportunity: methods that successfully achieve fine-grained alignment could substantially advance the field.

\subsection{Future Directions}

\subsubsection{Algorithm-Level Improvements}

A natural approach to leverage alignment would be to explicitly incorporate it into the optimization objective. For instance, the LSO algorithm could optimize both $R^2$ fitness and alignment quality (Equation~\eqref{eq:alignment}), guiding the search toward regions where candidate solutions are both numerically accurate and symbolically relevant to the target. However, our findings show this strategy is bottlenecked by alignment quality. Preliminary experiments confirm that while explicitly optimizing alignment increases $a_t$ during evolution, it does not improve symbolic retrieval performance when the underlying alignment is too coarse. This suggests that model-level improvements to alignment granularity are a prerequisite for effective algorithm-level exploitation.

\subsubsection{Model-Level Improvements}

Addressing coarse alignment in contrastive models is an active research area. Several techniques have been proposed to improve fine-grained alignment in CLIP and similar vision-language models \cite{lewis2022does,wang2023equivariant}. Adapting these techniques to mathematical expressions represents a promising direction. For symbolic regression, this might involve generating training data with carefully designed hard negatives (expressions that differ minimally but behave differently), incorporating structured contrastive objectives that explicitly reward fine-grained discrimination, or using auxiliary tasks that require detailed symbolic understanding.
Additionally, refined fine-tuning strategies after pretraining could improve alignment for specific SR applications. Such model-level improvements could then be combined with algorithm-level guidance mechanisms, enabling LSO procedures that truly exploit multi-modal information for symbolic search.

% \textcolor{blue}{Recent concurrent work supports this direction: GenSR \cite{li2026gensr} replaces contrastive pretraining entirely with a generative CVAE-based latent space, achieving strong results on SRBench. This suggests that the coarse-grained alignment we identify may be an inherent limitation of the contrastive (InfoNCE) objective for symbolic-numeric domains, rather than a problem solvable by fine-tuning alone. Exploring both refined contrastive objectives and alternative generative formulations appears warranted.}

\subsubsection{Broader Applications}

Beyond latent space optimization, improved multi-modal alignment could benefit symbolic regression in other ways. If alignment quality reaches sufficient granularity, it could serve as a learned symbolic similarity metric, offering advantages over hand-crafted measures like edit distance or tree distance. Such metrics could improve benchmarking methodologies and enable more principled evaluation of symbolic retrieval. Additionally, multi-modal representations could guide traditional GP operators, informing initialization, selection, or variation to maintain both numerical accuracy and symbolic coherence, similar to recent neural-guided GP methods \cite{anthes2025transformer,han2025transformer}.

%% file: sections/related_works.tex
\section{Related Work}

Building on Transformer architectures, several works propose one-shot inductive learning, mapping numerical observations directly to expressions to avoid iterative search \cite{biggio2021neural,kamienny2022end,vastl2024symformer,li2022transformer}. While offering fast inference, these models often struggle with exact symbolic pattern recovery on out-of-distribution data \cite{voigt2025analyzing,sato2025can,yu2024mdlformer}. % Recent improvements focusing on training data quality \cite{} and diffusion models \cite{} represent orthogonal directions that could eventually be combined with better alignment strategies.

 While one-shot models bypass iterative search entirely, other approaches use neural networks to enhance it. These methods maintain discrete symbolic representations but utilize neural networks or reinforcement learning to prioritize and guide traditional evolutionary search operators \cite{liskowski2018neuro,anthes2025transformer,han2025transformer,wyrwinski2025learning}. 

As an alternative to discrete search, LSO methods encode expressions into continuous spaces \cite{kusner2017grammar,dai2018syntax,meznar2023efficient,caetano2023symbolic}. While early approaches focused on modeling syntactic structure, our work specifically investigates SNIP \cite{meidani2024snip}, which attempts to bridge this gap using contrastive multi-modal learning. Concurrent to our study, GenSR \cite{li2026gensr} addresses the limitations of discriminative spaces by proposing a generative LSO approach via a dual-branch Conditional VAE. Crucially, their interpolation experiments independently confirm our findings that SNIP's contrastive space is highly fragmented. 

 Beyond continuous optimization, other recent frameworks recognize that SR inherently requires mapping between structural form and functional behavior. Methods like MMSR \cite{li2024mmsr} explicitly align symbolic and data-fitting objectives, while Bertschinger et al. \cite{bertschinger2024evolving} frame SR as a dual optimization of "form and function." Our fundamental investigation into fine-grained cross-modal alignment complements these works by highlighting the structural-behavioral bottlenecks present across SR paradigms.

%% file: sections/conclusion.tex
\section{Conclusion}

We investigated the promise of multi-modal learning for Genetic Programming, focusing on SNIP's application to symbolic regression. Through systematic experiments, we demonstrated two key limitations: (1) SNIP's latent space optimization does not actively exploit the learned cross-modal alignment during search, and (2) the alignment learned through contrastive pre-training is too coarse-grained to distinguish between structurally similar expressions. These findings explain the gap between SNIP's numerical success and its symbolic retrieval performance observed in prior work. While current bi-modal models do not yet achieve the fine-grained understanding required for principled symbolic search, our analysis clarifies the path forward: improving alignment granularity through refined training objectives, and designing optimization algorithms that explicitly leverage cross-modal relationships. We believe these directions can unlock the significant potential of multi-modal approaches for symbolic regression.